\documentclass[10pt]{article}
\usepackage[preprint]{tmlr}
\usepackage{microtype}
\usepackage{graphicx}
\usepackage{booktabs}
\usepackage[section]{placeins}
\usepackage{amsmath,amssymb,mathtools}
\usepackage{hyperref}
\usepackage{url}
\hypersetup{hidelinks,pdfauthor={C. Evans Hedges},pdftitle={The Inference-Compute Frontier for Limit Order Book Prediction}}
\newcommand{\R}{\mathbb{R}}
\newcommand{\ops}{\operatorname{ops}}
\title{The Inference-Compute Frontier\\for Limit Order Book Prediction}
\author{\name C. Evans Hedges\\
\addr Independent Researcher, New York, USA}

\begin{document}
\raggedbottom
\maketitle

\begin{abstract}
We identify an empirical inference-compute frontier for limit order book prediction on FI-2010. A power law fitted to low- and mid-compute models extrapolates to an excluded MLPLOB-style mixer family whose work is $56$--$2{,}515$ times the largest budget in the corresponding fit. Across nine folds and five horizons, pooled $R^2$ is $0.934$ on the realized target frontier, with 59\% lower RMSE than constant endpoint extrapolation, and $0.731$ on all excluded operating points. Measured latency gives a weaker relationship and reorders model families. We therefore design FastBiNLOB to pack computation into dense temporal and feature mixing. H120 exceeds the published state-of-the-art macro-F1 on two of four horizons in every one of five seeds. Under our CPU/MPS protocol, H96 reduces median latency by 23.6\% relative to the W128 MLPLOB reference, and H120 by 60.2\% relative to a TLOB-inspired dual-axis attention reference.
\end{abstract}

\section{Introduction}\label{sec:intro}

We study the best predictive loss attained at a given inference-compute budget in limit order book (LOB) prediction.\footnote{OpenAI Codex assisted with manuscript editing and consistency checks. The author takes responsibility for the final manuscript.} A forecast must be accurate enough to be useful and fast enough to act on. Improving accuracy by increasing inference work is therefore useful only if we can execute that work within the available time.

Power laws have described how predictive loss changes with training compute and model scale in other domains \citep{kaplan2020scaling,hoffmann2022chinchilla,edwards2024timeseriesscaling}. LOB prediction provides a different setting, with CNN/LSTM, attention, and dense MLP architectures \citep{zhang2019deeplob,tran2019tabl,berti2025tlob}. Here the object of study is the inference-compute frontier itself: the lower envelope of predictive loss across heterogeneous model families under a forward-work budget. We ask whether a power law fitted only to smaller classical and random-convolution models can locate the losses of an excluded neural family.

We study this question on FI-2010. For each fold and prediction horizon, we form the lower envelope of test cross-entropy against a structural count of forward work, then fit a power law to this envelope. We exclude the MLPLOB-style mixer family from fitting and evaluate it at three input windows whose compute is approximately $56\times$--$2{,}515\times$ beyond the fitted range. Across nine folds and five horizons, the fitted curves attain pooled $R^2=0.934$ on the realized mixer target frontier and $0.731$ on all 135 excluded operating points. On the target frontier, RMSE is 58.8\% lower than extrapolating the last non-mixer loss unchanged. The distinction between the two populations matters: a frontier concerns the best measured loss at each budget, and a larger model need not attain it. W256 illustrates this distinction, requiring more work without improving the target frontier.

We then consider what this relationship permits us to do under a latency constraint. Counted operations do not have a fixed execution time. Memory access, operator implementation, and the computational graph can make a model with more counted work run faster \citep{williams2009roofline,ma2018shufflenetv2,chen2018tvm}. Our timing results show this reordering directly. We therefore ask how to place useful computation in operations that execute quickly.

For this purpose, we introduce FastBiNLOB, a short stack of dense temporal and feature MLPs. H120 exceeds the published state-of-the-art macro-F1 on $y_{10}$ and $y_{100}$ by $0.51$ and $0.25$ percentage points on average, respectively \citep{berti2025tlob}; both improvements occur in all five seeds. FastBiNLOB also substantially reduces measured median latency: H96's median is 23.6\% lower than that of our W128 MLPLOB implementation, and H120's is 60.2\% lower than that of the TLOB-inspired dual-axis attention reference, using each model's faster CPU/MPS backend. FastBiNLOB returns all four horizon forecasts in one call. Section~\ref{sec:deployment_latency} specifies the reference graphs and timing protocol.

\section{Experimental setting}\label{sec:benchmark}

\subsection{Data and model families}

We use FI-2010 NoAuction Z-score data \citep{ntakaris2018benchmark}, which contain ten trading days from five Finnish stocks. Our scaling analysis uses the first 40 raw LOB features: bid/ask prices and volumes at ten levels. In chronological folds CF1--CF9, each fold trains on a prefix of days and tests on the next day. We use the supplied labels for horizons $y_{10},y_{20},y_{30},y_{50},y_{100}$, corresponding to the original labels $k\in\{1,2,3,5,10\}$. The task predicts downward, stationary, or upward mid-price movement, with loss
\begin{equation}\label{eq:loss}
L(f)=-\frac{1}{n}\sum_{i=1}^n\ln\hat p_f(y_i\mid x_i).
\end{equation}
All frontier losses are test cross-entropies in nats unless we explicitly refer to validation loss.

For the FastBiNLOB comparison, we use all 144 features, window $W=128$, train days 1--7, test days 8--10, and macro-F1 on $y_{10},y_{20},y_{50},y_{100}$. We compare accuracy with published MLPLOB/TLOB values for the same day split and feature set \citep{berti2025tlob}. We use the 40-feature input to compare a broad range of predictors and all 144 features to compare the proposed design with published LOB accuracy results. These choices affect benchmark comparisons \citep{prata2024lobbenchmark,briola2024lobframe}; the two settings do not define a single fitted frontier.

Our non-mixer fit pool contains 56 configurations in every fold/horizon cell: 9 decision trees, 30 histogram gradient boosting (HGB) models, 6 CatBoost symmetric-tree models, and 11 random-convolution logistic models, giving 2,520 evaluations. Discovery screened 833 CF1 evaluations covering 212 distinct window/model configurations. Selection used CF1 validation loss relative to the class-prior loss and retained coverage across compute budgets and windows. Six additional small W10 random-convolution models extend the low-compute coverage. We then keep the final set fixed across the subsequent cells. The accompanying source bundle contains the 56-configuration manifest, including windows and model-specific settings.

Random-convolution kernel lengths are drawn from $\{3,5,7,9,15,21,31,45,63\}$, restricted to lengths at most $W$. Each kernel acts on one randomly chosen input channel and contributes its maximum response, mean response, and fraction of positive responses. The logistic head uses $C=1$ and at most 800 iterations.

Our excluded neural family is an MLPLOB-style mixer, with one W64, W128, and W256 model per fold/horizon, giving 135 targets. Each uses SimpleBiN input normalization, a feature embedding with learned temporal bias, three fixed-width pre-LayerNorm residual blocks, final LayerNorm, temporal mean pooling, and a $40\to40\to3$ head. Hidden width is 40, expansion is four, and dropout is $0.1$. This family differs from the tapered published MLPLOB graph used as the deployment reference. Figure labels abbreviate the held-out family as MLPLOB.

Training uses AdamW, learning rate $3\times10^{-4}$, weight decay $10^{-4}$, batch sizes 32/64 for training/evaluation, at most 30 epochs, and patience eight. Minimum validation-loss improvement is $10^{-4}$ for W64/W256 and $5\times10^{-5}$ for W128. We reserve the final 20\% of training data for validation. Neural inputs receive per-lag/feature standardization using fitting-split statistics, applied unchanged to validation and test inputs before SimpleBiN. Each scaling operating point uses seed zero, and changing $W$ changes input history as well as compute.

\subsection{Counting inference work}\label{sec:work}

To compare these families, we need a compute convention that can be reconstructed from model metadata. We denote its one-observation count by $\ops(f)$. It counts selected forward operations, with one multiply-add counted as one unit, and omits memory traffic, vectorization, and framework dispatch. It is therefore a structural work proxy rather than a hardware instruction count.

For a tree with configured leaf budget $m$, we use $\log_2(m)+1$: a balanced-depth approximation and one leaf lookup. HGB with $I$ configured iterations uses $3I(\log_2(m)+2)$ for its three class-specific trees per iteration. CatBoost uses $I_{\mathrm{eff}}(d+1+3)$, with effective iteration count $I_{\mathrm{eff}}$, symmetric depth $d$, one vector-leaf lookup, and three logit additions.

For $K$ random kernels and valid lengths $S_W$, let $r_\ell=W-\ell+1$. We count temporal dot products, three pooled features per kernel, their standardization, and the linear classifier:
\begin{equation}\label{eq:randconv_work}
\ops_{\mathrm{RC}}=\frac{K}{|S_W|}\sum_{\ell\in S_W}(r_\ell\ell+4r_\ell-3)+15K.
\end{equation}
The average over lengths is an accounting expectation. For the MLPLOB-style mixer with $F$ inputs, width $H$, $L$ blocks, and expansion $E$, the dense count is
\begin{equation}\label{eq:mlplob_work}
C_{\mathrm{dense}}=WHF+L(2WH^2E+2HW^2E)+H^2+3H.
\end{equation}
The recorded W128 work uses this count. Recorded W64/W256 work also includes $24WH+H$ activation applications, a difference below 1\%. We retain these recorded coordinates for excluded-model scoring. Their values are 6,555,360, 20,850,360, and 73,402,080 for W64, W128, and W256, respectively.

\subsection{Fitting the frontier}\label{sec:frontier_fit}

For each fold and horizon, we first take the best measured loss at each available compute budget:
\begin{equation}\label{eq:frontier}
L_*(C)=\min\{L(f):f\in\mathcal M,\ \ops(f)\leq C\},
\end{equation}
where $\mathcal M$ is the non-mixer pool. To extrapolate this envelope, we fit
\begin{equation}\label{eq:power_law}
L(C)=L_\infty+AC^{-\alpha}
\end{equation}
to its strictly improving points by nonlinear least squares. Here $L_\infty$ is a nonnegative fitted offset, $A$ an amplitude, and $\alpha$ the decay exponent. Before fitting, we divide $C$ by its median on the envelope; this changes the amplitude's parameterization and improves numerical conditioning. The normalized fit bounds are $0\leq L_\infty<\min L$, $0\leq A\leq10$, and $0\leq\alpha\leq10$. No mixer-target loss enters the fit. We fit 45 separate curves, one per fold/horizon; pooled $R^2$ concatenates the excluded predictions before scoring against the overall target mean.

We score every excluded mixer model and, separately, its realized target frontier. The latter keeps a model when its observed loss improves on lower-work mixer models in the same cell, so its membership uses target outcomes. This evaluates the realized finite envelope, rather than using the fitted curve to select models during training. We report all nine folds and CF2--CF9. The latter restriction followed inspection of the smallest training fold, CF1, and is exploratory. The holdout is across architecture and compute; fitting and scoring still use the same benchmark test folds.

\section{The empirical power law}\label{sec:structural}

\begin{figure}[tbp]
\centering
\includegraphics[width=.82\linewidth]{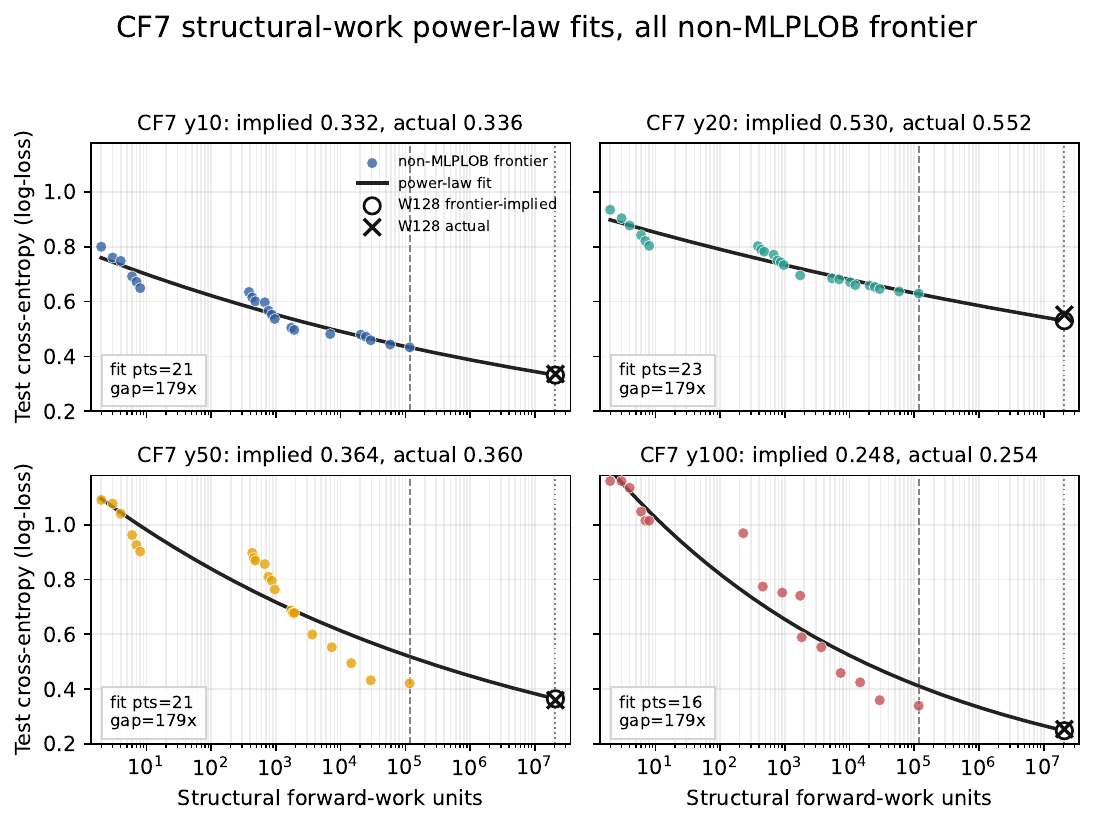}
\caption{Non-mixer frontier fits on four CF7 horizons. The excluded W128 models require substantially more compute; their predicted and observed losses are marked at the right.}
\label{fig:structural_frontier_cf7}
\end{figure}

\begin{figure}[tbp]
\centering
\includegraphics[width=.57\linewidth]{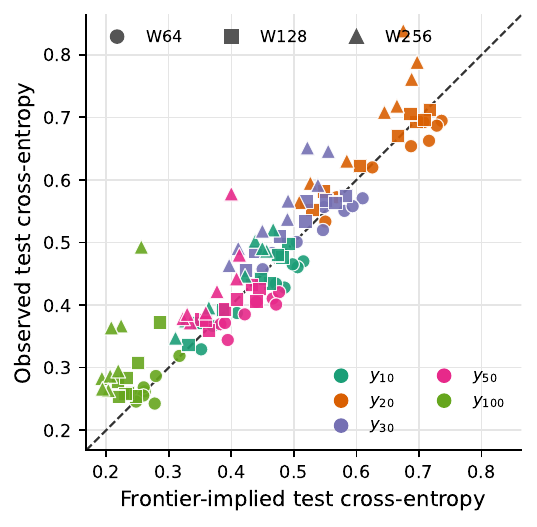}
\caption{Observed and predicted MLPLOB-style mixer test losses on CF2--CF9. Shapes distinguish windows and colors distinguish horizons. The largest window, W256, is consistently above the target frontier.}
\label{fig:mlplob_observed_predicted}
\end{figure}

The fitted non-mixer envelopes end between 29,184 and 116,736 work units. The mixer targets therefore test Equation~\ref{eq:power_law} well beyond the fitted compute range. Figure~\ref{fig:structural_frontier_cf7} shows example curves and excluded W128 losses, while Figure~\ref{fig:mlplob_observed_predicted} compares all excluded models on CF2--CF9.

To test whether extrapolating the curve adds information, we compare it with holding the last non-mixer frontier loss constant. We compute this endpoint prediction separately in every fold/horizon, without using mixer outcomes. Table~\ref{tab:endpoint} reports both methods on identical targets. On the realized target frontier across all nine folds, the power law reduces RMSE from $0.087$ to $0.036$, a 58.8\% reduction, and raises $R^2$ from $0.609$ to $0.934$. On CF2--CF9, frontier RMSE falls from $0.086$ to $0.032$, a 62.3\% reduction. Thus the fitted curve predicts continued improvement beyond the last low/mid-compute frontier point.

Across every excluded operating point, including dominated models, the complete-fold $R^2$ is $0.731$, and RMSE improves by about 5\% over the endpoint. On CF2--CF9 all targets, $R^2=0.838$ and RMSE falls from $0.077$ to $0.056$, about 27\%. In all 45 excluded-family test-loss fits, $L_\infty<10^{-10}$, so the observed scaling is carried by the decaying term.

\begin{table}[htbp]
\caption{Excluded mixer predictions from the power law (pow) and constant endpoint (end). Target-frontier membership is selected using observed target losses.}
\label{tab:endpoint}
\centering
% Generated by scripts/report_tmlr_existing_results.py; do not edit.
\begin{tabular}{llrrrrr}
\toprule
Folds & Targets & $n$ & $R^2_{\mathrm{pow}}$ & $R^2_{\mathrm{end}}$ & RMSE$_{\mathrm{pow}}$ & RMSE$_{\mathrm{end}}$\\
\midrule
CF1--CF9 & All targets & 135 & 0.731 & 0.703 & 0.082 & 0.086\\
CF1--CF9 & Target frontier & 47 & 0.934 & 0.609 & 0.036 & 0.087\\
CF2--CF9 & All targets & 120 & 0.838 & 0.693 & 0.056 & 0.077\\
CF2--CF9 & Target frontier & 42 & 0.941 & 0.581 & 0.032 & 0.086\\
\bottomrule
\end{tabular}

\end{table}

On CF2--CF9, W64 contributes all 40 of its points to the target frontier, W128 improves it in two cells, and W256 improves it in none. W64+W128 together give $R^2=0.948$, while W256 gives $0.647$. For W256, the endpoint prediction has lower RMSE ($0.066$) than the power law ($0.087$). We therefore see a regularity in the best attained losses, together with a concrete failure of a larger window to use its additional work effectively.

\begin{samepage}
The power law improves squared error over endpoint extrapolation on four of five horizons (Table~\ref{tab:horizon}). MSE skill, $1-\mathrm{SSE}_{\mathrm{pow}}/\mathrm{SSE}_{\mathrm{end}}$, ranges from $0.096$ to $0.902$ on those four tasks; positive values favor the power law. Its predictive value varies by horizon: within-horizon $R^2$ is near zero or negative on three horizons, and the endpoint performs better on $y_{100}$. Pooling tasks with different loss levels can increase $R^2$, so the horizon-specific errors complement the pooled frontier result. These are related tasks sharing data and training prefixes, rather than independent replications.

\end{samepage}

\begin{table}[htbp]
\caption{CF2--CF9 all-target errors by horizon: eight folds and three windows per row. Positive MSE skill favors the power law.}
\label{tab:horizon}
\centering
% Generated by scripts/report_tmlr_existing_results.py; do not edit.
\begin{tabular}{lrrrrrr}
\toprule
Horizon & $n$ & $R^2$ & MAE & RMSE & Endpoint RMSE & MSE skill\\
\midrule
$y_{10}$ & 24 & 0.634 & 0.028 & 0.034 & 0.108 & 0.902\\
$y_{20}$ & 24 & 0.490 & 0.039 & 0.053 & 0.082 & 0.579\\
$y_{30}$ & 24 & 0.007 & 0.041 & 0.051 & 0.062 & 0.322\\
$y_{50}$ & 24 & -0.304 & 0.039 & 0.053 & 0.056 & 0.096\\
$y_{100}$ & 24 & -1.017 & 0.055 & 0.079 & 0.066 & -0.449\\
\bottomrule
\end{tabular}

\end{table}

The cross-family relationship also survives substantial changes to the work accounting. In each of 500 draws, we multiply each family's work, including the neural mixer's, by an independent log-uniform factor in $[1/2,2]$, shared across all folds and horizons. We rebuild and refit the non-mixer envelopes while keeping the observed target-frontier membership fixed. On the CF1--CF9 target frontier, $R^2$ has mean $0.888$, median $0.900$, 5th--95th percentiles $0.770$--$0.951$, and minimum $0.698$. It remains at least $0.698$ in every draw, supporting stability to these family-level counting changes. The draw distribution measures accounting sensitivity rather than dataset sampling uncertainty.

The retrospective fit including mixer points on the full empirical frontier gives in-sample $R^2=0.902$ over 971 points. Within-cell $R^2$ has median $0.919$ and ranges from $0.726$ to $0.951$ across all 45 curves, so the fit describes curve shape within individual tasks as well as pooled loss levels.

Mean exponents range from $0.027$ on $y_{20}$ to $0.094$ on $y_{100}$. This retrospective analysis uses the dense count in Equation~\ref{eq:mlplob_work} for all mixer windows; its W64/W256 coordinates differ by less than 1\% from those used for excluded-model scoring.

The saved validation-to-validation analysis also favors the power law over the endpoint: all-target $R^2=0.667$ versus $0.612$ across all folds, and $0.809$ versus $0.565$ on CF2--CF9. The same ordering on this second partition complements the test-loss results. These descriptive checks support the measured relationship; the excluded-family results in Table~\ref{tab:endpoint} provide its extrapolation test.

\section{Using the frontier under a latency constraint}\label{sec:latency}

We now ask how to use the compute--loss relationship when the available resource is time. If additional computation can improve prediction, then a useful design goal is to execute that computation more efficiently. For this purpose, we first examine whether measured latency preserves the ordering supplied by structural work.

We replace work with configuration-level CPU timings, using representative CF7-fitted trees/HGB and instantiated inference paths for the other models. The configuration timing is reused across folds and horizons and includes library overhead. On the same 45 W128 targets across CF1--CF9, the structural power-law fit gives $R^2=0.893$. The best tested latency curve is exponential and gives $0.468$; the latency power law gives $-0.547$. The latency relationship is weaker even though its targets lie inside the fitted timing ranges, whereas the compute targets require extrapolation.

\begin{figure}[htbp]
\centering
\includegraphics[width=.72\linewidth]{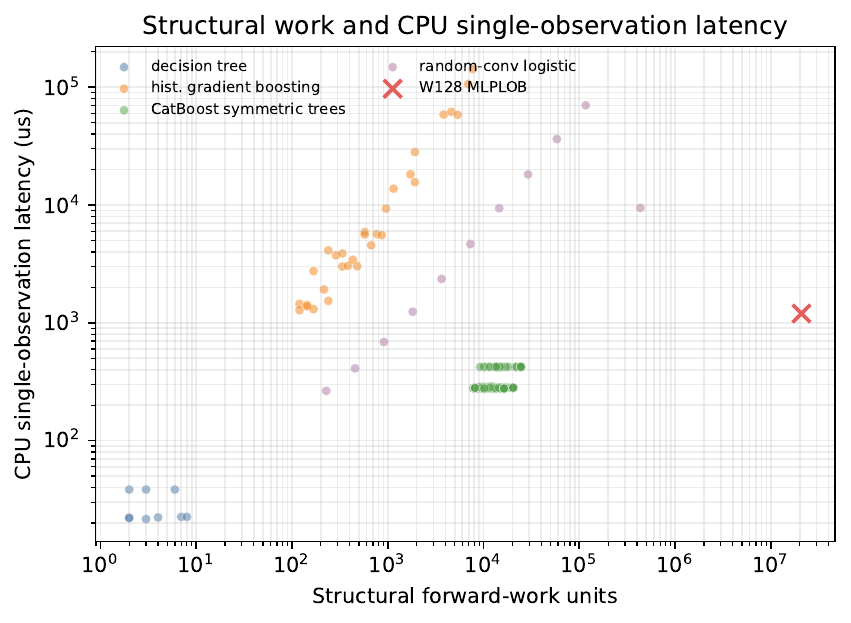}
\caption{Counted work and CPU single-observation latency for the 40-feature timing comparison. Each configuration is shown once. Families change order because timing depends on their graphs and software implementations.}
\label{fig:work_latency_scatter}
\end{figure}

Figure~\ref{fig:work_latency_scatter} shows the practical difference between the two resources. Median HGB work is 528 units with latency $4.34$ ms, while CatBoost has 12,276 units and $281\,\mu$s: approximately 23 times more counted work runs approximately 15 times faster under these implementations. The W128 mixer has 20.85 million units but runs in $1.20$ ms. Structural work and serving latency therefore supply different information for model design. More computation can fit into a shorter forward pass when its graph and software path execute it efficiently.

\FloatBarrier
\subsection{FastBiNLOB}\label{sec:fastbin}

FastBiNLOB combines BiN-style input normalization with temporal and feature MLPs, following the axis-separable mixing approach of all-MLP models \citep{tran2021bin,tolstikhin2021mlpmixer}. We begin with a 144-feature window $x\in\R^{144\times128}$. To accommodate differing feature scales and variation over time, SimpleBiN normalizes along each axis, learns a mixture of the results, and applies a featurewise affine map. We then transpose to time-major form, embed the features to width $H$, and add a learned temporal bias to obtain $z_0\in\R^{128\times H}$.

Each of two residual blocks applies a feature MLP $F_\theta$ independently at every time step, followed by a temporal MLP $G_\phi$ independently on each channel:
\begin{equation}\label{eq:mixer}
z'=z+F_\theta(z),\qquad z_{\mathrm{out}}=z'+G_\phi((z')^\top)^\top.
\end{equation}
This gives global temporal mixing through fixed learned matrices, with nonlinear feature updates. Like MLPLOB, the graph is attention-free; compared with attention-based TLOB, it omits pairwise attention maps, Q/K/V projections, and attention softmax. It trades content-adaptive pairwise interactions for a short sequence of dense linear operations.

Both variants use feature expansion one, GELU, dropout $0.05$, no block/final LayerNorm, and four output heads, each with one hidden layer of width $H$ and three logits. H96 mean has $H=96$, temporal expansion four, and mean pooling. H120 taper has $H=120$ and temporal expansion ten. Its reducer maps channels $H\to2H\to H/4$, then time $128\to256\to32$, before flattening for the heads. The models have 365,581 and 1,289,555 parameters, respectively.

\subsection{Accuracy and training}\label{sec:deployment_training}

We train jointly on four horizons with equal weights and logit-adjusted cross-entropy ($\tau=0.5$) \citep{menon2021logit}. For horizon $h$ and class $c$, training adds $0.5\log\pi_{h,c}$ to the class logit, using add-one-smoothed fit-set priors $\pi_{h,c}$; inference uses the unadjusted logits. We use Adam at $3\times10^{-4}$, zero weight decay, $\epsilon=10^{-8}$, training/evaluation batch sizes 32/64, and an exponential moving average with decay $0.999$. Selection maximizes mean validation macro-F1 on the final 20\% of training data, with at most 50 epochs, patience 15, and minimum improvement $10^{-5}$. Windows respect day boundaries. Lagged inputs are standardized coordinatewise using fit-partition means and standard deviations, then the same transformation is applied to validation and test data. We use seeds 1305--1309.

Table~\ref{tab:fastbin_macro_f1} reports one jointly selected EMA checkpoint per seed, with all four horizons evaluated together and no horizon-specific checkpoint selection, fine-tuning, or class-bias calibration. Values are mean $\pm$ sample standard deviation over seeds; the mean over horizons is computed within each seed first. We take reference accuracy from \citet{berti2025tlob}, whose \href{https://github.com/LeonardoBerti00/TLOB/blob/main/models/engine.py}{public implementation} reports macro-F1. Their best reported windows are W384 for MLPLOB and W128 for TLOB. The local timing graphs, specified below, are separate from these published accuracy results.

\begin{table}[htbp]
\caption{Macro-F1 with all 144 features. FastBiNLOB values are five-seed mean $\pm$ sample standard deviation; reference accuracy is published. The mean covers the four displayed horizons.}
\label{tab:fastbin_macro_f1}
\centering
\setlength{\tabcolsep}{3pt}
% Generated by scripts/report_tmlr_existing_results.py; do not edit.
\begin{tabular}{lrrrrr}
\toprule
Model & Mean & $y_{10}$ & $y_{20}$ & $y_{50}$ & $y_{100}$\\
\midrule
Published MLPLOB & .8763 & .8164 & .8488 & .9139 & .9262\\
Published TLOB & .8677 & .8155 & .8268 & .9003 & .9281\\
H96 mean & $0.8647\pm0.0003$ & $0.8184\pm0.0006$ & $0.8121\pm0.0011$ & $0.9004\pm0.0003$ & $0.9278\pm0.0014$\\
H120 taper & $0.8678\pm0.0008$ & $0.8215\pm0.0011$ & $0.8158\pm0.0013$ & $0.9035\pm0.0006$ & $0.9306\pm0.0005$\\
\bottomrule
\end{tabular}

\end{table}

H120 exceeds the published state-of-the-art macro-F1 on $y_{10}$ (MLPLOB) and $y_{100}$ (TLOB) by $0.51$ and $0.25$ percentage points on average, respectively. Both improvements occur in every seed. H96 also exceeds the state-of-the-art $y_{10}$ reference in every seed. H120's four-horizon mean is $0.8678$, below MLPLOB's $0.8763$, with lower scores on $y_{20}$ and $y_{50}$. These are consistent seed-level improvements on the two named horizons; the single published reference scores do not provide a matched significance test.

\subsection{Packing compute into the forward pass}\label{sec:deployment_latency}

We time 10,000 batch-one forward-plus-softmax calls after 256 warmup rows from CF8 test inputs. The model and inputs are resident on the device; preprocessing and transfer are excluded. Models run in evaluation and inference mode. MPS is synchronized immediately before and after each timed call. We instantiate fixed-shape graphs rather than loading each selected seed checkpoint. FastBiNLOB produces four heads in one call; each reference graph produces one horizon.

The MLPLOB reference is a device-neutral port of the \href{https://github.com/LeonardoBerti00/TLOB/blob/main/models/mlplob.py}{published implementation}, with H144/L3 at $W=128$. The TLOB-inspired reference uses H16/L1, four attention heads, and feed-forward multiplier two at $W=128$. It embeds each scalar input into a width-16 token, applies temporal attention per input feature and spatial attention per time step, and mean-pools both axes. This graph differs from the \href{https://github.com/LeonardoBerti00/TLOB/blob/main/models/tlob.py}{published TLOB implementation}, which embeds whole feature vectors and reduces the two axes before a dense head. Its timings therefore characterize the stated dual-axis attention reference, while the accuracy comparison uses published TLOB scores.

We use a MacBook Pro (Mac15,8), Apple M3 Max with 16 CPU cores and 128 GB RAM, macOS 26.5.1, Python 3.12.2, and PyTorch 2.12.0 in eager FP32. CPU execution uses 12 intra-op and 16 inter-op threads; MPS uses the Apple GPU backend. Calls are separated by 1 ms. Table~\ref{tab:fastbin_latency} reports the resulting percentiles.

\begin{table}[htbp]
\caption{Batch-one forward-plus-softmax latency in $\mu$s, under the stated CPU/MPS protocol. TLOB-inspired H16 denotes the dual-axis attention reference described in the text. FastBiNLOB returns four horizon forecasts per call; each reference returns one. Percentiles are from timed calls, not independent benchmark runs.}
\label{tab:fastbin_latency}
\centering
% Generated by scripts/report_tmlr_existing_results.py; do not edit.
\begin{tabular}{llrrrr}
\toprule
Model & Backend & P50 & P90 & P95 & P99\\
\midrule
MLPLOB & CPU & 2101 & 2400 & 2509 & 2794\\
MLPLOB & MPS & 2834 & 3252 & 3369 & 3728\\
TLOB-inspired H16 & CPU & 22846 & 24727 & 25248 & 25897\\
TLOB-inspired H16 & MPS & 5089 & 7194 & 7725 & 8883\\
H96 mean & CPU & 1604 & 2002 & 2199 & 2985\\
H96 mean & MPS & 2634 & 3150 & 3371 & 3970\\
H120 taper & CPU & 2028 & 2599 & 2813 & 3266\\
H120 taper & MPS & 2124 & 2310 & 2396 & 2898\\
\bottomrule
\end{tabular}

\end{table}

H120 uses almost three times H96's counted work (94.53 versus 31.69 million operations), with a CPU median of 2,028 versus 1,604 $\mu$s and higher mean macro-F1 on all four horizons. H120 also executes more work than MLPLOB's 91.33 million operations in a slightly shorter median CPU call (2,028 versus 2,101 $\mu$s). In this setting, dense mixing lets us fit more computation into a short forward pass.

H96's CPU median is 23.6\% lower than MLPLOB's. Choosing each model's lower-P50 backend gives the TLOB-inspired reference on MPS at $5{,}089\,\mu$s and H120 on CPU at $2{,}028\,\mu$s, a 60.2\% reduction. These per-call gains are obtained while FastBiNLOB returns all four horizon forecasts jointly. The improvement depends on the percentile: H96's CPU P99 is higher than MLPLOB's (2,985 versus 2,794 $\mu$s), and H120 has higher CPU P90--P99 despite its 3.5\% lower median. The table retains both backends and all reported percentiles.

The 144-feature experiment applies the design reasoning suggested by the 40-feature frontier; it does not place FastBiNLOB on that fitted curve. We compare complete training recipes and one hardware/software protocol, so the accuracy differences cannot be attributed to architecture alone, nor the timings treated as hardware-independent speedups. Together, the results establish a useful design distinction on this benchmark: an empirical frontier describes the predictive losses attained at larger work budgets, while serving measurements show how efficiently a concrete graph executes that work. FastBiNLOB turns this distinction into a model with consistent gains over the two horizon-specific SOTA scores and lower measured median latency under the stated comparisons.

\clearpage
\bibliography{tmlr_references}

@article{ntakaris2018benchmark,
  title={Benchmark Dataset for Mid-Price Forecasting of Limit Order Book Data with Machine Learning Methods},
  author={Ntakaris, Adamantios and Magris, Martin and Kanniainen, Juho and Gabbouj, Moncef and Iosifidis, Alexandros},
  journal={Journal of Forecasting},
  volume={37},
  number={8},
  pages={852--866},
  year={2018},
  doi={10.1002/for.2543},
  eprint={1705.03233},
  archivePrefix={arXiv}
}

@article{zhang2019deeplob,
  title={{DeepLOB}: Deep Convolutional Neural Networks for Limit Order Books},
  author={Zhang, Zihao and Zohren, Stefan and Roberts, Stephen},
  journal={IEEE Transactions on Signal Processing},
  volume={67},
  number={11},
  pages={3001--3012},
  year={2019},
  eprint={1808.03668},
  archivePrefix={arXiv}
}

@article{tran2019tabl,
  title={Temporal Attention Augmented Bilinear Network for Financial Time-Series Data Analysis},
  author={Tran, Dat Thanh and Iosifidis, Alexandros and Kanniainen, Juho and Gabbouj, Moncef},
  journal={IEEE Transactions on Neural Networks and Learning Systems},
  volume={30},
  number={5},
  pages={1407--1418},
  year={2019},
  doi={10.1109/TNNLS.2018.2869225}
}

@article{tran2021bin,
  title={Bilinear Input Normalization for Neural Networks in Financial Forecasting},
  author={Tran, Dat Thanh and Kanniainen, Juho and Gabbouj, Moncef and Iosifidis, Alexandros},
  journal={arXiv preprint arXiv:2109.00983},
  year={2021},
  eprint={2109.00983},
  archivePrefix={arXiv}
}

@article{prata2024lobbenchmark,
  title={{LOB}-based Deep Learning Models for Stock Price Trend Prediction: A Benchmark Study},
  author={Prata, Matteo and Masi, Giuseppe and Berti, Leonardo and Arrigoni, Viviana and Coletta, Andrea and Cannistraci, Irene and Vyetrenko, Svitlana and Velardi, Paola and Bartolini, Novella},
  journal={Artificial Intelligence Review},
  volume={57},
  pages={116},
  year={2024},
  doi={10.1007/s10462-024-10715-4}
}

@article{briola2024lobframe,
  title={Deep Limit Order Book Forecasting},
  author={Briola, Antonio and Bartolucci, Silvia and Aste, Tomaso},
  journal={arXiv preprint arXiv:2403.09267},
  year={2024},
  eprint={2403.09267},
  archivePrefix={arXiv}
}

@article{berti2025tlob,
  title={{TLOB}: A Novel Transformer Model with Dual Attention for Price Trend Prediction with Limit Order Book Data},
  author={Berti, Leonardo and Kasneci, Gjergji},
  journal={arXiv preprint arXiv:2502.15757},
  year={2025},
  eprint={2502.15757},
  archivePrefix={arXiv}
}

@article{kaplan2020scaling,
  title={Scaling Laws for Neural Language Models},
  author={Kaplan, Jared and McCandlish, Sam and Henighan, Tom and Brown, Tom B. and Chess, Benjamin and Child, Rewon and Gray, Scott and Radford, Alec and Wu, Jeffrey and Amodei, Dario},
  journal={arXiv preprint arXiv:2001.08361},
  year={2020},
  eprint={2001.08361},
  archivePrefix={arXiv}
}

@article{hoffmann2022chinchilla,
  title={Training Compute-Optimal Large Language Models},
  author={Hoffmann, Jordan and Borgeaud, Sebastian and Mensch, Arthur and Buchatskaya, Elena and Cai, Trevor and Rutherford, Eliza and de Las Casas, Diego and Hendricks, Lisa Anne and Welbl, Johannes and Clark, Aidan and others},
  journal={arXiv preprint arXiv:2203.15556},
  year={2022},
  eprint={2203.15556},
  archivePrefix={arXiv}
}

@article{edwards2024timeseriesscaling,
  title={Scaling-laws for Large Time-series Models},
  author={Edwards, Thomas D. P. and Alvey, James and Alsing, Justin and Nguyen, Nam H. and Wandelt, Benjamin D.},
  journal={arXiv preprint arXiv:2405.13867},
  year={2024},
  eprint={2405.13867},
  archivePrefix={arXiv}
}

@inproceedings{ma2018shufflenetv2,
  title={{ShuffleNet V2}: Practical Guidelines for Efficient {CNN} Architecture Design},
  author={Ma, Ningning and Zhang, Xiangyu and Zheng, Hai-Tao and Sun, Jian},
  booktitle={Proceedings of the European Conference on Computer Vision},
  pages={116--131},
  year={2018}
}

@article{williams2009roofline,
  title={Roofline: An Insightful Visual Performance Model for Multicore Architectures},
  author={Williams, Samuel and Waterman, Andrew and Patterson, David},
  journal={Communications of the ACM},
  volume={52},
  number={4},
  pages={65--76},
  year={2009},
  doi={10.1145/1498765.1498785}
}

@inproceedings{chen2018tvm,
  title={{TVM}: An Automated End-to-End Optimizing Compiler for Deep Learning},
  author={Chen, Tianqi and Moreau, Thierry and Jiang, Ziheng and Zheng, Lianmin and Yan, Eddie and Shen, Haichen and Cowan, Meghan and Wang, Leyuan and Hu, Yuwei and Ceze, Luis and Guestrin, Carlos and Krishnamurthy, Arvind},
  booktitle={13th USENIX Symposium on Operating Systems Design and Implementation},
  pages={578--594},
  year={2018}
}

@article{tolstikhin2021mlpmixer,
  title={{MLP-Mixer}: An all-{MLP} Architecture for Vision},
  author={Tolstikhin, Ilya and Houlsby, Neil and Kolesnikov, Alexander and Beyer, Lucas and Zhai, Xiaohua and Unterthiner, Thomas and Yung, Jessica and Steiner, Andreas and Keysers, Daniel and Uszkoreit, Jakob and Lucic, Mario and Dosovitskiy, Alexey},
  journal={Advances in Neural Information Processing Systems},
  volume={34},
  year={2021}
}

@inproceedings{menon2021logit,
  title={Long-Tail Learning via Logit Adjustment},
  author={Menon, Aditya Krishna and Jayasumana, Sadeep and Rawat, Ankit Singh and Jain, Himanshu and Veit, Andreas and Kumar, Sanjiv},
  booktitle={International Conference on Learning Representations},
  year={2021},
  eprint={2007.07314},
  archivePrefix={arXiv}
}
\bibliographystyle{tmlr}
\end{document}